\documentclass[conference]{IEEEtran}

\usepackage{cite}
\usepackage[cmex10]{amsmath}
\usepackage{graphicx}

\usepackage[hyphens]{url}

\begin{document}

\title{Semantic Image Retrieval \\ via Active Grounding of Visual Situations}
\author{\IEEEauthorblockN{Max H. Quinn$^1$, Erik Conser$^1$, Jordan M. Witte$^1$, and Melanie Mitchell$^{1,2}$}
\IEEEauthorblockA{$^1$Portland State University  $^2$Santa Fe Institute \\ Email: quinn.max@gmail.com}}

\maketitle

\begin{abstract}
 
We describe a novel architecture for semantic image retrieval---in
particular, retrieval of instances of {\it visual situations}.  Visual
situations are concepts such as ``a boxing match,'' ``walking the
dog,'' ``a crowd waiting for a bus,'' or ``a game of ping-pong,''
whose instantiations in images are linked more by their common spatial
and semantic structure than by low-level visual similarity.  Given a
query situation description, our architecture---called {\it
  Situate}---learns models capturing the visual features of expected
objects as well the expected spatial configuration of relationships
among objects.  Given a new image, Situate uses these models in an
attempt to {\it ground} (i.e., to create a bounding box locating)
each expected component of the situation in the image via an active
search procedure.  Situate uses the resulting grounding to compute a
score indicating the degree to which the new image is judged to
contain an instance of the situation.  Such scores can be used to rank
images in a collection as part of a retrieval system.  In the
preliminary study described here, we demonstrate the promise of this
system by comparing Situate's performance with that of two baseline
methods, as well as with a related semantic image-retrieval system
based on ``scene graphs.''

\end{abstract}

\section{Introduction}

The ability to automatically retrieve images with specified semantic
properties is a key topic for computer vision.  In a world deluged
with image data, automated image retrieval has become as important as
text search, and progress in this area will have profound impacts in
areas as diverse as medical diagnosis, public health, national
security, privacy, and personal data organization.

Using deep neural networks, automatic detection of individual objects
in images has become remarkably successful \cite{Ren2015}.  However,
in many domains, users need to search for images with more abstract
properties, in which multiple objects with specified attributes are
related in specific ways.  Here are some examples: ``a boxing match,''
``a person walking a dog,'' ``a crowd waiting for a bus,'' or ``a game
of ping-pong.''  Instances of such abstract visual concepts---which we
call {\it visual situations}---are linked more by their common spatial
and semantic structure than by low-level visual similarity or by the
specific objects they contain.  In general, automatically recognizing
instances of a given visual situation is a difficult problem, due to
substantial variability in visual features and spatial layout among
different instances.  Moreover, while state-of-the-art object
detection methods often rely on evaluating large numbers of ``object
proposals'' at every location and scale of the image, the
combinatorics of such exhaustive evaluations become much worse when
the multiple objects, attributes, and possible relationships of a
situation need to be considered.  And while successful object
detection has relied on huge amounts of labeled training data
\cite{Socher2009}, there are few large labeled training sets for
visual situations.

\begin{figure*}[t]
\centering
\includegraphics[width=6.5in]{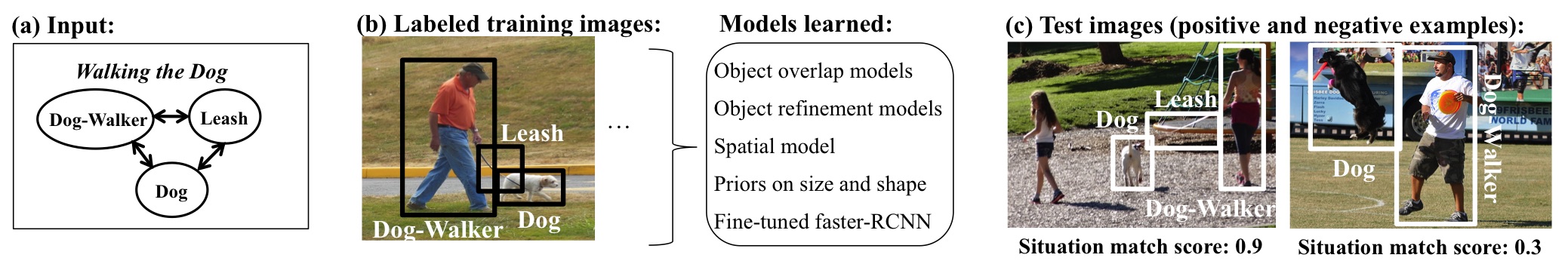}
\caption{(Best viewed in color.) Overview of Situate's training and
  testing pipeline, with the example situation ``walking the
  dog.''  (a) The user specifies the object categories relevant to the
  given situation.  This simple example includes three expected
  objects: {\it Dog-Walker}, {\it Leash}, and {\it Dog}; the
  double-arrows indicate unnamed (i.e., to be learned) relationships
  among these objects.  (b) Situate is also given a set of training
  images, which include only positive examples of the situation, with
  relevant objects indicated via human-labeled (black) bounding boxes.  From these
  training examples, Situate learns several types of models.  (c)
  Situate uses its learned models to score test images as instances of
  the given situation, by attempting to ground the expected object
  categories and their expected spatial configuration in the image (white boxes).
}
\label{SituatePipeline}
\end{figure*}

In this paper, we describe a novel architecture for retrieving
instances of a query visual situation in a collection of images.  Our
architecture---called {\it Situate}---combines object-localization
models based on visual features with probabilistic models that
represent learned multi-object relationships.  Situate
learns these models from labeled training images.  It applies these
models to a new image via an active search process that attempts to
ground components of the query situation in the image---that is, to
create bounding boxes that localize relevant objects and
relationships, and that ultimately provide a {\it situation match
  score} for the situation with respect to the image. The match scores
can be used to rank the images in the collection with respect to the
query: the highest ranking images can be returned to the user.
Figure~\ref{SituatePipeline} illustrates Situate's training and
testing pipeline.

We hypothesize that Situate's learned object and relationship models, used
in tandem with its active situation-grounding method, will result in
superior image retrieval performance than methods without these
components.  In this preliminary study, we test this hypothesis by
comparing Situate's performance on a challenging visual-situation
dataset with that of two baseline methods: a ``lesioned'' version of
Situate that lacks its relationship models and the feedback they provide
the system, and an adapted version of the widely used Faster-RCNN
object-detection method \cite{Ren2015}.  We also compare Situate's
performance with that of a recently proposed semantic image retrieval system
\cite{Johnson2015a}.

\section{Situate's Architecture \label{SituateArchitecture}}

Situate's architecture is inspired by {\it active} approaches to
perception, in which the perceiver acquires information dynamically,
and in which the information acquired continually feeds back to
control the perceptual process.  In particular, for humans,
recognizing a visual situation is an active process that unfolds over
time, in which prior knowledge interacts with visual information as it
is perceived, in order to guide subsequent eye movements and focus of
attention.  This interaction of top-down expectations and bottom-up
perception enables a human viewer to very quickly locate relevant
aspects of the situation \cite{Neider2006,Summerfield2009}.

Additionally inspired by the {\it Copycat} architecture of Hofstadter
and Mitchell \cite{Hofstadter1994}, Situate attempts to actively
ground components of a query situation via the actions of
{\it agents} in a {\it Workspace}.  The agents are selected and run
over a series of time steps, and create candidate detections by
combining bottom-up visual information with top-down expectations.
The advantage of this active, temporal approach is that detections
made in previous time steps can influence the focus and actions of
agents at subsequent time steps.  

\subsection{Training \label{training}}

In attempting to ground situation components in an image, Situate's
agents dynamically combine different types of information by employing
a set of models that capture expected visual and spatial features of
the query situtation.  These models are learned from human-labeled
training data, illustrated in Figure~\ref{SituatePipeline}(b).  
The learned models are the following:

{\bf Object-Localization Models:} For a given situation query, Situate
learns object-localization models for each relevant object category.  The
input to each object-localization model is an {\it object proposal}, which
specifies an object category $C$ along with a set of coordinates
specifying a bounding box $b$ in an image.  For an object-localization
model corresponding to category {\it C}, the model's output is a
prediction of the amount {\it b} overlaps a ground-truth
object of category {\it C}.  Following standard practice in the
object-detection literature, overlap is measured as the {\it
  intersection over union} (IOU) of the proposed bounding box with a
(human-labeled) ground-truth bounding box.

We implemented each object-localization model as a linear combination of
features, where the features are obtained from running a pretrained
convolutional network on the input region.  We used the open-source
VGG-f network pre-trained on Imagenet \cite{Imagenet-vgg} to obtain
4096 features from the fc7 layer, and used ridge regression
\cite{MatlabRidgeRegression} to learn the coefficients and bias of the
linear model.  We trained the ridge regression model on
features extracted from training-image crops that partially or
completely overlap ground-truth boxes.

{\bf Object-Refinement Models:} Situate similarly learns
category-specific {\it object-refinement models} from the same
training crops used to learn object-localization models.  These
refinement models---based on the ``bounding-box regression'' approach
described in \cite{Girshick2014}---input an object proposal and output
a new, ``refined'' object proposal that is predicted to have higher
IOU with a ground-truth bounding box of the given category.  Like the object-localization models described
above, each category-specific object-refinement model is a linear
combination of 4096 features from the pre-trained VGG fc7 layer.
As we will describe below, on a new test image the refinement models
will be applied to object proposals whose predicted overlap score is
above a pre-set threshold.  We found these refinement models to be
crucial to Situate's ability to localize objects.

{\bf Relationship Model:} For a given situation query, Situate learns a
{\it relationship model} that represents the expected spatial relationships
among the relevant objects in that situation.  The relationship model is a
multivariate Gaussian distribution learned from human-labeled bounding
boxes in training images.  The variables in the distribution are
bounding-box parameters---center coordinates, area ratio (i.e, area of
box divided by area of image), and aspect ratio---from the relevant
objects in a given situation.  For example, for the {\it Walking the
  Dog} situation shown in Figure~\ref{SituatePipeline}(a), the
variables are the bounding-box parameters of the {\it Dog-Walker}, {\it Dog}, and {\it Leash}
boxes.  As we will describe below, when Situate is run on a new image,
each time it makes a candidate detection of one or more relevant
objects, it conditions the relationship model on those detections to narrow
the expected location, shape, and size of the related objects.

{\bf Priors on Object Size and Shape:} Situate also learns a model
capturing prior expectations of each relevant object category's size
(area ratio) and shape (aspect ratio). These expectations are learned
by fitting the area ratios and aspect ratios of ground-truth boxes as
independent category-specific log-normal distributions.  Log-normals
are a better fit for the distribution of these values than normal
distributions because the former are always positive and give more
weight to smaller values.  Note that our system does not learn prior
distributions over bounding-box {\it location}, since we do not want
the system to model photographers' biases to put relevant objects near
the center of the image.

{\bf Fine-Tuned Faster-RCNN:} Before starting its run on a test image,
Situate runs a fine-tuned version of faster-RCNN on the test image to
create a ``prior'' set of potential object proposals (to be described
in the next section).  Faster-RCNN \cite{Ren2015} is a widely-used
deep convolutional network that is trained to propose bounding boxes
and score them with respect to given object categories; it has been
shown to achieve state-of-the-art performance on object detection.  We
used an open-source version of Faster-RCNN \cite{Girshick2015a} that
was pre-trained on the Pascal VOC dataset, and we used Situate's
training set to fine-tune it for the {\it Dog-Walker}, {\it Dog}, and
{\it Leash} categories. 

\subsection{Running Situate on a Test Image}

After the models described above have been learned from training data,
Situate is ready to run on new (``test'') images.  The input to
Situate is an image and the program's output is (1) a {\it situation
  match score} that measures Situate's assessment of this image as an
instance of the given situation, and (2) a set of
``groundings''---detections of situation components in the Workspace,
as illustrated in Figure~\ref{SituatePipeline}(c).

\begin{figure*}[t]
\centering
\includegraphics[width=6.8in]{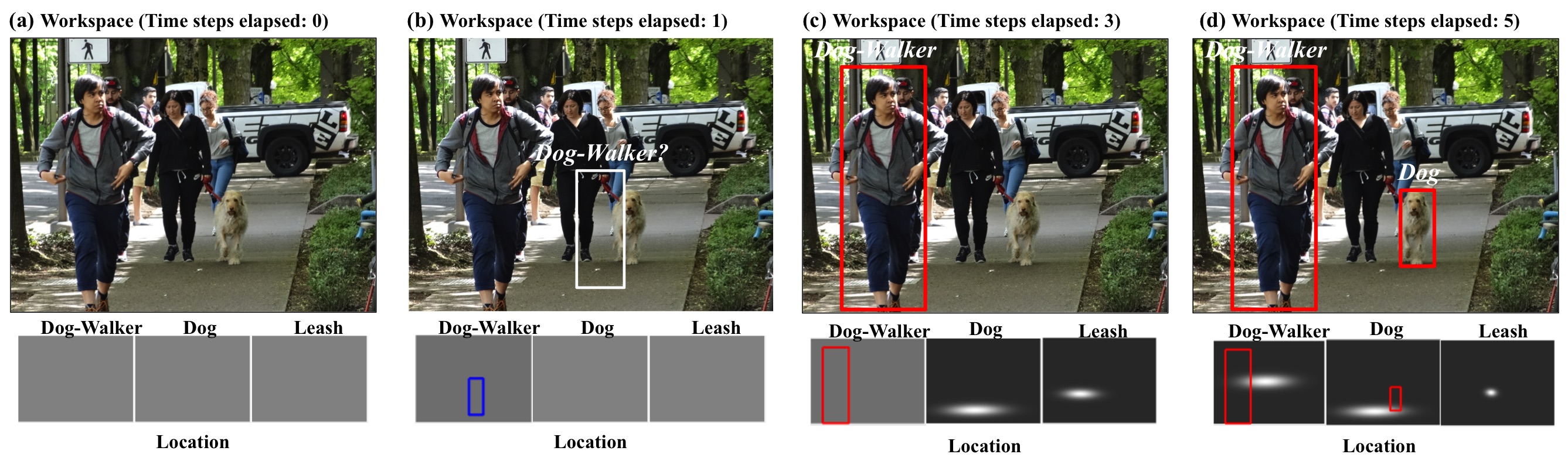}
\vspace*{.05in}
\includegraphics[width=6.8in]{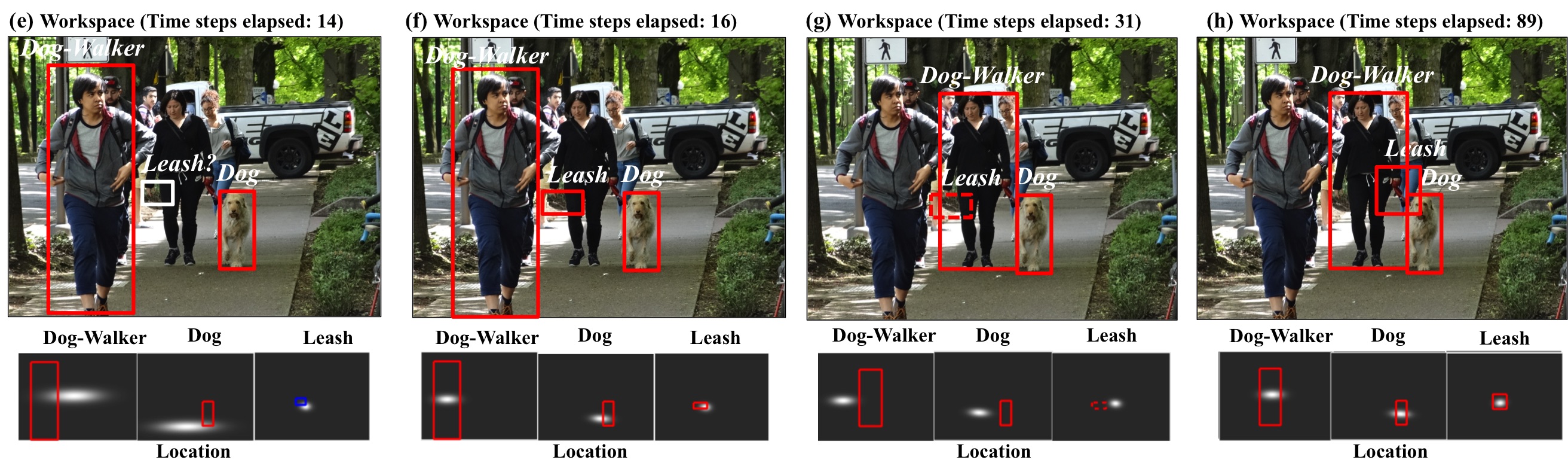}
\caption{(Best viewed in color.) Visualization of eight time steps in a run of Situate.  See text for explanation. }
\label{SituateRun}
\end{figure*}

The detailed process by which Situate runs on a test image is
illustrated in Figure~\ref{SituateRun}, which shows visualizations of
eight time-slices from a run of the program using the {\it Walking the
  Dog} situation of Figure~\ref{SituatePipeline}(a).

{\bf Figure~\ref{SituateRun}(a):} The figure shows Situate's state before any
agents have run.  The Workspace contains the unprocessed image.
The gray squares shown below the Workspace represent the probability
distributions over location for each object category.  The uniform
gray indicates that these are initially uniform distributions.  Once
an object is detected, these distributions will be updated to be
conditioned on that detection according to the learned relationship model.
As we described above, the system also maintains
probability distributions (not shown here) for aspect ratio
(``shape'') and area ratio (``size'') of each object category.
Initially these shape and size distributions are set to the learned
independent priors for each category, but once an object is detected
these distributions will also be conditioned on that detection.

As we described above, Situate will attempt to ground each relevant
object (here, {\it Dog-Walker}, {\it Dog}, and {\it Leash}) in the
Workspace via the action of agents.  The system maintains an
{\it agent pool}---a collection of agents waiting to run---and selects
one of them at random to run at each time step.

In order to balance exploration for new object proposals with
``exploitation'' and refinement of known, promising object proposals,
Situate has three types of agents: {\it explorers}, {\it refiners},
and {\it RCNN-priors}.  An explorer agent chooses an object category
from the list of relevant categories (here, {\it Dog-Walker}, {\it
  Dog}, and {\it Leash}), and samples from that category's current
location, shape, and size distributions to create an object proposal.
A refiner attempts to improve an already-existing object proposal.  An
{\it RCNN-prior} agent proposes one of the bounding boxes that were
computed prior to Situate's run by running our fine-tuned faster-RCNN
on the test image.  We found that all three of these types of agents
were needed in order for Situate to work well.  The importance of
explorer agents will be seen in Section~\ref{results} when we compare
Situate's performance with that of faster-RCNN alone.

The agent pool is initialized with $P$ RCNN-prior agents for each
object category---these represent the $P$ highest-confidence boxes for
each category, as computed and scored by faster-RCNN---and $P'$
explorer agents.  In the experiments described in
Section~\ref{methods}, we set $P=10$ and $P'=30$---that is, 10
RCNN-priors per category, and 30 explorer agents (each of which will choose
a category at random when it runs).  These values were chosen via experiments on a validation subset of
the training images.

Once an agent runs, it is permanently deleted from the pool.  Every
time an explorer agent is run and deleted, a new explorer agent is
added to the pool, in addition to any follow-up refiner agents that
were added by previously run agents.

The challenging aspects of this particular test image are (1) to identify the correct person as
the dog-walker and (2) to locate the small, hard-to-see leash.

{\bf Figure~\ref{SituateRun}(b):} At time step 1 an explorer agent is
selected from the pool to run.  It chooses the {\it Dog-Walker}
category, and samples from the location, size, and shape distributions
of that category to produce a proposal, represented as a white box in
the Workspace labeled ``{\it Dog-Walker?}'' (and a corresponding blue
box in the location distribution).  The explorer agent evaluates this
proposal via two measures: {\it internal} and {\it external} support.
The internal support is the {\it Dog-Walker} object-localization model's
prediction of overlap between this proposal and a ground-truth
dog-walker.  The external support is a function that uses the learned
relationship model to measure how well this proposal fits in with other
detections that have been made (none so far in this example).  These
two measures are combined into a {\it total support} measure, which
reflects the system's current judgement of the quality of this
proposal for the given situation.  If the internal support is greater
than a pre-defined threshold, the proposal will be marked for 
refinement; if the total support is above a second threshold, the
proposal will be accepted as a {\it detection}.  Since internal
support is very low for this particular {\it Dog-Walker} proposal, it
will be discarded without followup.  (Due to space constraints in this paper, the
specific forms of the internal, external, and total support functions,
along with values of the thresholds we used, will be given in an
online supplementary information section when this paper is
published.)

{\bf Figure~\ref{SituateRun}(c):} At time step 3 an RCNN-prior agent
is selected to run.  It creates a {\it Dog-Walker} proposal (in fact,
this is faster-RCNN's most confident ``Dog-Walker'' box for this
image).  This proposal's internal support is high enough for the
system to create a {\it detection} (red box in the Workspace;
corresponding red box in the location distribution).  A detection is a
data structure in the Workspace indicating that the system is
confident that it has located a particular object.  The relationship model
is now conditioned on this detection, yielding new location
distributions for {\it Dog} and {\it Leash}, shown in the boxes below
the Workspace.  (The size and shape distributions for these
categories, not shown here, are also conditioned on the detection.)
Agents will now increasingly focus on searching for objects in
higher-probability areas of these distributions (shown as white regions in the
location distributions).  Note that the {\it Dog-Walker} location
distribution is still uniform, since no dogs or leashes have been
detected yet.  The only problem is that the program has identified the
wrong person as the dog-walker.  This will be corrected in subsequent
time steps using additional information discovered by the program.

{\bf Figure~\ref{SituateRun}(d):} At time step 5 an RCNN-prior agent
creates a {\it Dog} detection.  The location distributions (as
well as size and shape distributions) for each category are now
conditioned on the detections of the other categories.   In particular, notice
that the {\it Dog-Walker} location distribution has been conditioned
on the {\it Dog} detection, and that the current {\it Dog-Walker}
detection is offset from the center of that distribution.  Similarly,
the new {\it Dog} detection is offset from the center of the {\it Dog}
location distribution, which is conditioned on the {\it Dog-Walker}.
In short, these detections do not support each other strongly.  Even
though both {\it Dog-Walker} and {\it Dog} have been detected, these
(and likewise any detection) are treated as provisional until the end
of the run: agents will continue to search for better-fitting
alternatives.

{\bf Figure~\ref{SituateRun}(e):} At time step 14, an explorer
agent samples from {\it Leash} distributions to produce a proposal
(white box).  While the internal strength of this proposal is not high
enough for it to be a {\it detection}, a refiner agent is created to
improve it, and is added to the agent pool. 

{\bf Figure~\ref{SituateRun}(f):} The refiner agent runs at
time step 16, creating a leash detection.  (False-positive leash
detections are very commonly made, both by faster-RCNN and by
Situate.)  The location distributions for each category are updated to
reflect this new detection.  While the {\it Dog-Walker} and {\it
  Leash} detections strongly support each other, the {\it Dog}
detection does not fit in as well.

{\bf Figure~\ref{SituateRun}(g):} An explorer agent has proposed an
alternative {\it Dog-Walker}, and due to its high internal support as
well as external support from the {\it Dog} detection, this new {\it
  Dog-Walker} has higher total support than the previous {\it
  Dog-Walker} detection, and has replaced it.  The relationship model
has been updated to reflect the new set of detections.  This causes
the existing {\it Leash} detection to lose much of its external (and
thus total) support; its weakness is illustrated by the dashed line
around it.

{\bf Figure~\ref{SituateRun}(h):} A stronger {\it Leash} detection has
replaced the weaker one, and all three detections now strongly support one another
(each can be seen to be at the center of its location distribution).
This mutual context has helped the system identify the correct
dog-walker as well as to locate the small, hard-to-see leash.  The run
ends at this point, since all three objects have been detected.  (If
not all objects are detected before a pre-set maximum
number of time steps, the run stops.)

At the end of a run, the system computes the {\it
  situation match score} as a function of the total support of each
detection.  In the current version of Situate, we define the situation
match score as the geometric mean over the ``padded'' total support
values of detections in the Workspace, where we ``pad'' the total
support values by adding 0.01 to each in order to avoid multiplication
by zero.  If no detection for a given object is made by the end of the
run, the situation match score is set to the minimum value of 0.01.
We chose this padded geometric mean function as a simple way of
combining total support scores, but will investigate alternative
scoring methods in future work.

In summary, the following sketches the initialization and main loop of Situate. 
\begin{list}{}{}
\item {\bf Input:} A test image. 
\item {\bf Initialization:} Initialize location, area-ratio, and
  aspect-ratio distributions for each relevant object category.  The
  initial location distributions are uniform; initial area-ratio and
  aspect-ratio distributions are learned from training data.
  Initialize agent pool with explorer and RCNN-prior agents. 
\item {\bf Main Loop:} Repeat until all relevant objects are detected or at most for {\it Max-Iterations}: 
\begin{enumerate}
     \item Choose agent from agent pool at random. 
     \item Run agent and delete it from the agent pool (and if agent
       is an explorer, replace it in the agent pool).
     \item Update relationship model, conditioned on current detections in the Workspace. 
\end{enumerate}
\item {\bf Return:} Situation match score $S$, where
\begin{equation}
S = \left[\prod_{i=1}^{n} \left( \text{total-support}(d_i) + 0.01\right) \right]^\frac{1}{n}, 
\end{equation}
where $n$ is the number of detections in the Workspace, and $d_i$ is the $i$th detection.  
\end{list}
For the experiments described in this paper, we used {\it
  Max-Iterations} = 300.  That is, at most, 300 agents are run
including explorers, refiners, and RCNN-priors.

In designing Situate's architecture, we were inspired by Hofstadter et
al.'s idea of modeling perception as a ``parallel terraced scan''
\cite{Hofstadter1994}, in which many possible exploration paths are
pursued in parallel, but not all at the same speed or to the same
depth.  Moreover, the exploration is ``active'' in the sense that
information is used as it is gained to continually modify the
resources given to possible paths.  An advantage of such an approach
is balancing the need to explore many possibilities while still
avoiding exhaustive evaluation of possible situation configurations.
Like the {\it codelets} in the architecture of \cite{Hofstadter1994},
our (serially run) architecture approximates such a parallel search
strategy by interleaving many independent agents.  In principle, many
of these explorations could be performed in parallel.  Furthermore,
splitting up ``explorers'' and ``refiners'' allows the system to
balance time spent on bottom-up exploration with time on focused
follow-ups.

Due to space constraints, we omit some details of Situate (e.g., the
detailed forms of the external and total support functions, thresholds
for detections) and the other methods.  We will provide these details
in an online supplementary materials section, along with our code and
all training and test data, upon publication of this paper.

We hypothesize that the approach we have described above will have
superior performance on grounding elements of situations, and thus on
ranking images, than methods that do not use this kind of active
approach, assuming the same amount of training data.  We also
hypothesize that our method will be able to achieve this performance
with significantly fewer object-proposal evaluations than non-active
methods.

\section{Related Work}

Here we describe some of the recent approaches most closely related to
Situate's goals and architecture. Closely related to our work is the
approach of Johnson et al. \cite{Johnson2015a} for semantic image
retrieval via ``scene graphs.''  We describe this method in Section~\ref{methods} and compare its performance to that of Situate in Section~\ref{results}. 

Another widely studied image-understanding task is that of ``grounding
referential expressions'' (e.g.,
\cite{Mao2016,Nagaraja2016,Rohrbach2016}).  Given a phrase such as
``the brown dog next to the woman wearing sunglasses,'' the task is to
locate the object being referred to, by grounding each object and
relation in the phrase.   Like the scene-graph task described above, research on this task
has focused on specific ``free-form'' phrases rather than more
abstract situation descriptions. The open-ended nature of the task
makes it very difficult, and accuracies reported on large datasets
have remained low to date.  A related task is that of detecting visual
relationships in images (e.g., \cite{Lu2016}); to our knowledge, the
literature on this task has focused almost exclusively on pairwise
relationships (e.g. ``dog riding surfboard''), rather than
multi-object visual situations.

Our situation-retrieval task shares motivation but contrasts with the
well-known tasks of ``event recognition'' or ''action recognition'' in
still images (e.g., \cite{Guo2014,Li2007,Wang2015}).  These latter
tasks consist of classifying images into one of several event or
action categories, without the requirement of localizing objects or
relationships.  A related task, dubbed ``Situation Recognition'' in
\cite{Yatskar2016}, requires a system to, given an image, predict the
most salient verb, along with its subject and object (``semantic
roles'' \cite{Gupta2015}).

Our task also contrasts with recent work on automatic caption
generation for images (e.g., \cite{Vinyals2015}), in which image
content is statistically associated with a language generator.  The
goal of caption-generation systems is to generate a description of
{\it any} input image.  Even the versions with ``attention'' (e.g., \cite{Xu2015}),
which are able to highlight diffuse areas corresponding roughly to
relevant objects, are not generally able to recognize and locate all important
objects, relationships, and actions, or more generally to recognize
abstract situations.  

While the literature cited above does not include {\it active}
detection methods such as Situate that involve feedback, there has
been also considerable work on active object detection (e.g.,
\cite{Alexe2012,Gonzalez-Garcia2015}), often in the context of active
perception in robots \cite{Ballard1991} and modeling visual attention
\cite{Ba2014,Mnih2014}.  More recently, several groups have
framed active object detection as a Markov decision process and use
reinforcement learning to learn a search policy (e.g., \cite{Caicedo2015}).

This section has given a sampling of recent work most closely related
to Situate.  While our work shares motivation with some of these
efforts, the specific problem we are addressing (visual situation
retrieval) and method (active grounding of situation components) is,
to our knowledge, unique in the literature.

\section{Datasets \label{Dataset}}

The computer vision community has created several important benchmark
datasets for object recognition and detection (e.g.,
\cite{Everingham2010,Socher2009}) and for some of the other tasks
described in the previous section that combine vision and language
(e.g., \cite{Krishna2017,Lin2014}).  None of these offers
the kind of data that we needed for our situation-retrieval
task---that is, collections of numerous instances of specific
multi-object situations, in which the objects are localized with
ground-truth bounding boxes. (For example, the ImSitu ``Situation
Recognition'' dataset is organized around verbs such as {\it
  carrying}, {\it jumping}, and {\it attacking}, each with one subject
(e.g., ``dog jumping'') and some with one additional object that the
subject acts upon (e.g., ``man carrying baby'').

For our preliminary work with Situate, we developed a new dataset
representing the ``Walking the Dog'' situation.  We chose this
situation category because it is reasonably easy to find sufficient
varied instances to train and test our system, and these instances
offer a variety of interesting challenges.  This dataset, the
``Portland State Dog-Walking Images,'' contains 500 positive instances
of the {\it Dog-Walking} situation.  These positive instances are
photographs taken by members of our group, and in each we labeled
(with bounding boxes) the {\it Dog}, {\it Dog-Walker}, and {\it
  Leash}.  Each image contains only one of each target object, but
many also contain additional (non-dog-walking) people, along with
cars, buildings, trees, and other ``clutter.''  The challenges of this
dataset include determining which person is the dog-walker, as well as
locating dogs (often small, and sometimes partially occluded) and
leashes (which are very often difficult, based on visual features
alone, to distinguish from other line-like structures in an image),
and deciding if the configuration of these three objects fits the
learned dog-walking situation.  For the experiments described below,
we split the 500 images into a 400-image training set and a 100-image
test.   We also created a negative set of 400 images selected from the Visual
Genome dataset \cite{Krishna2017}, including images in which people
interact with dogs in non-dog-walking situations, along with images
with people but no dogs, dogs but no people, and neither.  

\begin{table*}[t]
\caption{Results for Single-Image Recall$@N$.  Best results for each $N$ are in boldface.}  
\label{RecallAtNTable}
\centering
\begin{tabular}{|l|l|l|l|l|l|l|} \hline
N           & 1          & 2      & 5          & 10         & 20         & 100  \\ \hline
Situate     & {\bf 0.37} (0.09) & {\bf 0.50} (0.07)         & {\bf 0.56} (0.07) & {\bf 0.65} (0.04) & {\bf 0.77} (0.05) & {\bf 0.93} (0.02)  \\ \hline
Uniform     & 0.29 (0.07))& 0.32 (0.07)   & 0.44 (0.06)   & 0.54 (0.04)      & 0.65 (0.03)       & 0.79 (0.03) \\ \hline
Faster-RCNN & 0.24      & 0.25     & 0.28       & 0.38       & 0.54       & 0.91 \\ \hline
IRSG        & 0.24      & 0.24     & 0.27       & 0.37       & 0.55       & 0.87 \\ \hline
\end{tabular}

\end{table*}

\section{Methods \label{methods}}

We performed experiments to evaluate Situate's image retrieval and
situation grounding abilities.  We assess the importance of Situate's
learned relationship models by comparing with two baseline methods as
well as with the Image Retrieval using Scene Graphs (IRSG) method of
Johnson et al. \cite{Johnson2015a}.  In each method, we performed any
necessary training using the same training set that we used for
Situate.  All methods are tested on the ``Portland State Dog Walking''
test set.

{\bf Baseline Methods:} The first baseline, which we call the {\it
  Uniform} method, is identical to Situate except that explorer agents
always sample locations and bounding box parameters uniformly rather
than using a learned relationship model.  When creating an object
proposal, an explorer agent chooses a center location by sampling
uniformly across the entire image, and chooses area and aspect ratios
by sampling uniformly over fixed ranges.  The second baseline uses our
fine-tuned version of Faster-RCNN (described in Section~\ref{training}
above).  We ran the fine-tuned Faster-RCNN network on each test image
(positive and negative), and selected the highest scoring bounding box
(as scored by Faster-RCNN) corresponding to each of the three relevant
object categories.  Analogous to Situate, we defined the {\it
  Situation Match Score} as the padded geometric mean of the scores
assigned to these bounding boxes by Faster-RCNN.  Our goal was to see
how well this specially trained Faster-RCNN model could be used to
perform situation retrieval by simply relying on the top-scoring boxes
for each relevant object, absent of any relationship model or active
search.

{\bf IRSG Method:} We also compare Situate's performance that of
Johnson et al.'s ``Image Retrieval using Scene Graphs'' (IRSG) method
\cite{Johnson2015a}.  A scene graph is a graphical representation of
objects, attributes, and relationships that encode an image or image
region, or desired image content---e.g., ``a tall man wearing a white
baseball cap.''  IRSG is similar to Situate in that it scores an input
image as to how well it instantiates a query description (represented
as a scene graph), and uses these scores to rank images in a
collection.  Moreover, IRSG computes its score by attempting to ground
components of the query scene graph in the input image.  IRSG first
runs the geodesic object proposal method of \cite{Krahenbuhl2014} to
create a set of non-category-specific object proposal bounding boxes
(on the order of several hundred per image).  IRSG then uses R-CNN
\cite{Girshick2014} to give each bounding box multiple ``appearance''
scores---one for each object category in the scene graph.  Then, for
each possible pair of bounding boxes, IRSG uses a set of Gaussian
mixture models (GMM), learned from training examples, to score the
pair on each relationship in the scene graph .  The {\it unary} object
appearance scores and {\it binary} relationship scores are used in a
conditional random field model defined over a factor graph
representing the query.  As a simple example, for the query ``woman
next to man wearing hat,'' the system would give an appearance score
to each bounding box for ``woman,'' ``man,'' and ``hat,'' and then
give a relationship score to each pair of bounding boxes for each of
the two relationships (``man wearing hat'' and ``woman next to man''),
and see which configuration of object boxes and relationships
minimizes the energy function defined by the conditional random field.

We obtained the source code for IRSG from the authors of
\cite{Johnson2015a} and adapted it in order to compare it with Situate
and our other methods.  Instead of geodesic object proposals scored by
R-CNN, we used the top-scoring 300 boxes per category ({\it
  Dog-Walker}, {\it Dog}, {\it Leash}) from our fine-tuned version of
Faster-RCNN.  Since IRSG is (as currently implemented) limited to
pairwise relationships, we trained GMMs to represent three spatial
relationships: {\it Dog-Walker} and {\it Leash}; {\it Leash} and {\it Dog}; and {\it Dog-Walker}
and {\it Dog}.  The conditional random field formulation and energy
minimization was performed using the same algorithms that were
described in \cite{Johnson2015a}.  In this way, each positive and
negative test image in our set was scored with its final energy value,
and images were ranked in order of increasing energy---i.e., the
lowest energy image was ranked the highest.  (In the other methods,
images were ranked in order of decreasing score---the highest scoring
image was ranked the highest.)

{\bf Evaluation Metric:} We ran each method on the 100 positive test images and the the 400
negative (non-dog-walking) test images chosen from the Visual Genome
dataset.  Our evaluation metric is Single-Image Recall $@ N$
(abbreviated $R@N$): the probability that, if a single positive
example were added to the set of negative examples, and the collection
was ranked by situation match score (or energy, for IRSG), the
positive example would be in the $N$ top-ranked images.  For example,
given our 100 positive and 400 negative test images, $R@10=0.65$ means
that 65 out of 100 of the positive images would be in the top 10
ranked images if they were ranked alone with the 400 negative images.

\section{Results \label{results}}

Table~\ref{RecallAtNTable} gives the $R@N$ values resulting from
running the four different methods on the positive and negative test sets. 
While Faster-RCNN and our adapted version of IRSG are deterministic,
the Situate and Uniform methods are stochastic.  We performed 10
independent runs of each of these latter methods; each run consisted of running on all
the positive and negative test images, and then computing $R@N$.  The
values given in Table~\ref{RecallAtNTable} for Situate and Uniform are
the averages over these $R@N$ values (with standard deviations in
parentheses).  It can be seen that Situate produced the highest $R@N$ of
any method for each value of $N$; for low values of $N$, which are
typically the most important for retrieval tasks, Situate's $R@N$ was
substantially higher than any of the other methods.  These results support
our hypothesis that Situate's active grounding method,
together with its learned models, will result in superior image
retrieval performance than methods lacking these components.

\section{Discussion}

We investigated the reasons for for Situate's superior performance by
viewing the object detections produced by each method by the end of a
run on each test image.  As was illustrated by the run in Figure~\ref{SituateRun}, Situate was
considerably better at identifying the correct person as {\it
  Dog-Walker} and at locating hard-to-see leashes than the other
methods.  Moreover, Situate was better at locating small or partially
occluded objects than the other methods.  Also, in the case of
faster-RCNN, even when it did locate a hard-to-see object, faster-RCNN often
gave it low confidence, making the situation-match score low.  In
contrast, due to its external support measure, Situate was able to
assign higher confidence to such objects and thus score positive
situation instances more highly than faster-RCNN.

All methods were susceptible to false-positive object detections with
high confidence, but again, because of its incorporation of external
support for object detections, Situate was less susceptible than RCNN
and IRSG to giving high situation match scores to negative images.  In
principle IRSG should have also given higher energy (lower score) to
such images due to its relationship models, but it was not very
effective at doing so.  We will investigate the reasons for its failures in detail
in the near future.

While we have reported results on only one particular situation query
(``Dog-Walking''), we believe that the mechanisms that lead to
Situate's superior retrieval performance are general and
scalable. Demonstrating this scalability and generality is the most
important topic of our near-term future work.  

\section{Conclusions and Future Work}

In this paper we have described a preliminary study of Situate, a
novel approach to semantic image retrieval.  The results of this study
have shown the promise of Situate's active situation-grounding
architecture: our system's image-retrieval performance on the
``Walking the Dog'' situation strongly surpassed that of two baselines
as well as a related image-retrieval system from the recent
literature.  We showed how Situate is able to use information as it is
gained in order to focus its search, and to use the support of context
in order to locate hard-to-detect objects (e.g., barely visible
leashes, small dogs, partially occluded objects).  In analyzing these
results, we were able to understand some of the reasons for Situate's
superior performance, as well as to identify some of its problems.
This analysis underscores the important role of {\it grounding}
situation elements as part of scoring an image.

Visual situation recognition and retrieval is a broad and difficult
open problem in computer vision research, and the results we have
presented highlight many avenues of future research.  In the near term
we plan to improve our algorithm in several ways: 
expanding the kind of object attributes that can be detected by agents
(e.g., orientation and other pose features); expanding the types of
relationships that can be identified (e.g., recognizing that two
objects have the same orientation).  In the work described here we
used a multivariate Gaussian to capture spatial relationships among
objects; this simple probabilistic model is very fast to learn and to
sample from.  We plan to experiment with more sophisticated relationship
models
while keeping in
mind the tradeoff between sophistication and speed of computation.
Most importantly, we will explore the ability of our algorithms to
scale to larger datasets and to generalize to other situation
categories.

In the longer term, we will focus on, among other extensions, being
able to speed up our active search method via parallelization.
Finally, one of our original motivations for this project was to
create a system that can recognize visual {\it analogies}.  For
example, most people would consider images of a person running, riding a bicycle, sitting in a wheelchair, etc. while ``walking'' a dog to still be instances
of the abstract {\it Dog-Walking} situation.  This kind of recognition
requires what Hofstadter and colleagues have called ``conceptual
slippage,'' in which the roles defining a situation can be fluidly
filled by concepts semantically related to the prototype. Making
appropriate conceptual slippages is at the heart of analogy-making,
which itself is a core aspect of cognition \cite{Hofstadter1994}.

The abilities of computer vision remain far from human-level visual
understanding, but we believe that progress on the problem of
situation recognition, particularly incorporating analogy-making, will
play a pivotal role in giving computers the ability to make sense of
what they see.


\section*{Acknowledgments}
We are grateful to Justin Johnson of Stanford University for sharing
the source code for the IRSG project, and to NVIDIA corporation for
donation of a GPU used for this work.  Many thanks to Garrett Kenyon,
Bryan Lee, Sheng Lundquist, Anthony Rhodes, Evan Roche, Rory Soiffer,
and Robin Tan for discussions and assistance concerning this
project. This material is based upon work supported by the National
Science Foundation under Grant Number IIS-1423651.  Any opinions,
findings, and conclusions or recommendations expressed in this
material are those of the authors and do not necessarily reflect the
views of the National Science Foundation.

\bibliography{Situate}
\bibliographystyle{IEEEtran}

\end{document}